\documentclass[final]{cvpr}

\usepackage{times}
\usepackage{epsfig}
\usepackage{graphicx}
\usepackage{amsmath}
\usepackage{amssymb}
\usepackage{comment}
\usepackage{multirow}

\usepackage{color}

\usepackage[pagebackref=true,breaklinks=true,colorlinks,bookmarks=false]{hyperref}

\newcommand{\tablestyle}[2]{\setlength{\tabcolsep}{#1}\renewcommand{\arraystretch}{#2}\centering\small}
\newlength\savewidth\newcommand\shline{\noalign{\global\savewidth\arrayrulewidth
  \global\arrayrulewidth 1pt}\hline\noalign{\global\arrayrulewidth\savewidth}}

\def\cvprPaperID{****} 
\def\confYear{CVPR 2021}

\begin{document}

\title{Proposal Relation Network for Temporal Action Detection}

\author{
Xiang Wang$^{1,2}$
\quad Zhiwu Qing$^{1,2}$ 
\quad Ziyuan Huang$^{2}$ 
\quad Yutong Feng$^{2}$ 
\quad Shiwei Zhang$^{2*}$
\\
\quad Jianwen Jiang$^2$ 
\quad Mingqian Tang$^2$ 
\quad Changxin Gao$^1$
\quad Nong Sang$^{1*}$ 
\\
$^1$ Key Laboratory of Image Processing and Intelligent Control \\
School of Artificial Intelligence and Automation, Huazhong University of Science and Technology\\
$^2$Alibaba Group\\
{\tt\small \{wxiang, qzw, cgao, nsang\}@hust.edu.cn}\\
{\tt\small \{pishi.hzy, fengyutong.fyt, zhangjin.zsw, jianwen.jjw, mingqian.tmq\}@alibaba-inc.com}
}

\maketitle

\let\thefootnote\relax\footnotetext{$*$ Corresponding authors.}
\let\thefootnote\relax\footnotetext{This work is supported by Alibaba Group through Alibaba Research Intern Program.}
\let\thefootnote\relax\footnotetext{This work is done when X. Wang (Huazhong University of Science and Technology), Z. Qing (Huazhong University of Science and Technology), Z. Huang (National University of Singapore) and Y. Feng (Tsinghua University) are interns at Alibaba Group.}

\begin{abstract}
This technical report presents our solution for temporal action detection task in AcitivityNet Challenge 2021.
The purpose of this task is to locate and identify actions of interest in long untrimmed videos.
The crucial challenge of the task comes from that the temporal duration of action varies dramatically, and the target actions are typically embedded in a background of irrelevant activities.
Our solution builds on BMN~\cite{bmn}, and mainly contains three steps:
1) action classification and feature encoding by Slowfast~\cite{slowfast}, CSN~\cite{csn} and ViViT~\cite{VIVIT};
2) proposal generation. We improve BMN by embedding the proposed Proposal Relation Network (PRN), by which we can generate proposals of high quality;
3) action detection. We calculate the detection results by assigning the proposals with corresponding classification results.
Finally, we ensemble the results under different settings and achieve \textcolor{blue}{\textbf{44.7\%}} on the test set, which improves the champion result in ActivityNet 2020~\cite{wang2020cbr} by \textcolor{blue}{\textbf{1.9\%}} in terms of average mAP.
\end{abstract}

\section{Introduction}

Action understanding is an important area in computer vision, and it draws growing attentions from both industry and academia because its use in human computer interaction, public security and some other far reaching applications.
It includes many sub-research directions, such as Action Recognition~\cite{tsn,slowfast,huang2021self}, Temporal Action Detection~\cite{bmn,qing2021temporal,wang2021self}, Spatio-Temporal Action Detection~\cite{song2019tacnet,anetava2018}, etc. 
In this report, we introduce our method for the temporal action detection task in the $6$-th ActivityNet challenge~\cite{caba2015activitynet}.

For temporal action detection task, we need to localize and classify the target actions simultaneously.
Current mainstream approaches~\cite{bmn,gtad,tsp} are designed in a two-stage pipeline, \emph{i.e.}, proposal generation and action classification, and have achieved remarkable performance.
Therefore, we follow this paradigm to design the solution of this challenge.
Moreover, to further improve the performance, we design a proposal relation module to capture the relations among the proposals by non-local operations, and thus can also better model long temporal relationships.
%

%
%

%

\section{Action Recognition and Feature Extraction}
In recent years, a large number of deep networks for action recognition are proposed~\cite{csn,slowfast,i3d}. 
They have been playing an important role in the development of action detection.

\subsection{Slowfast}
Slowfast network~\cite{slowfast} was proposed for action classification by combining a fast and a slow branch. 
For the slow branch, the input is with a low frame rate, which is used to capture spatial semantic information. 
The fast branch, whose input is with a high frame rate, targets to capture motion information. 
Note that the fast branch is a lightweight network, because its channel is relatively small. 
Due to its excellent performance in action recognition and detection, we choose Slowfast as one of our backbone models.

\begin{figure*}[t!]
\centering
\centering{\includegraphics[width=.8\linewidth]{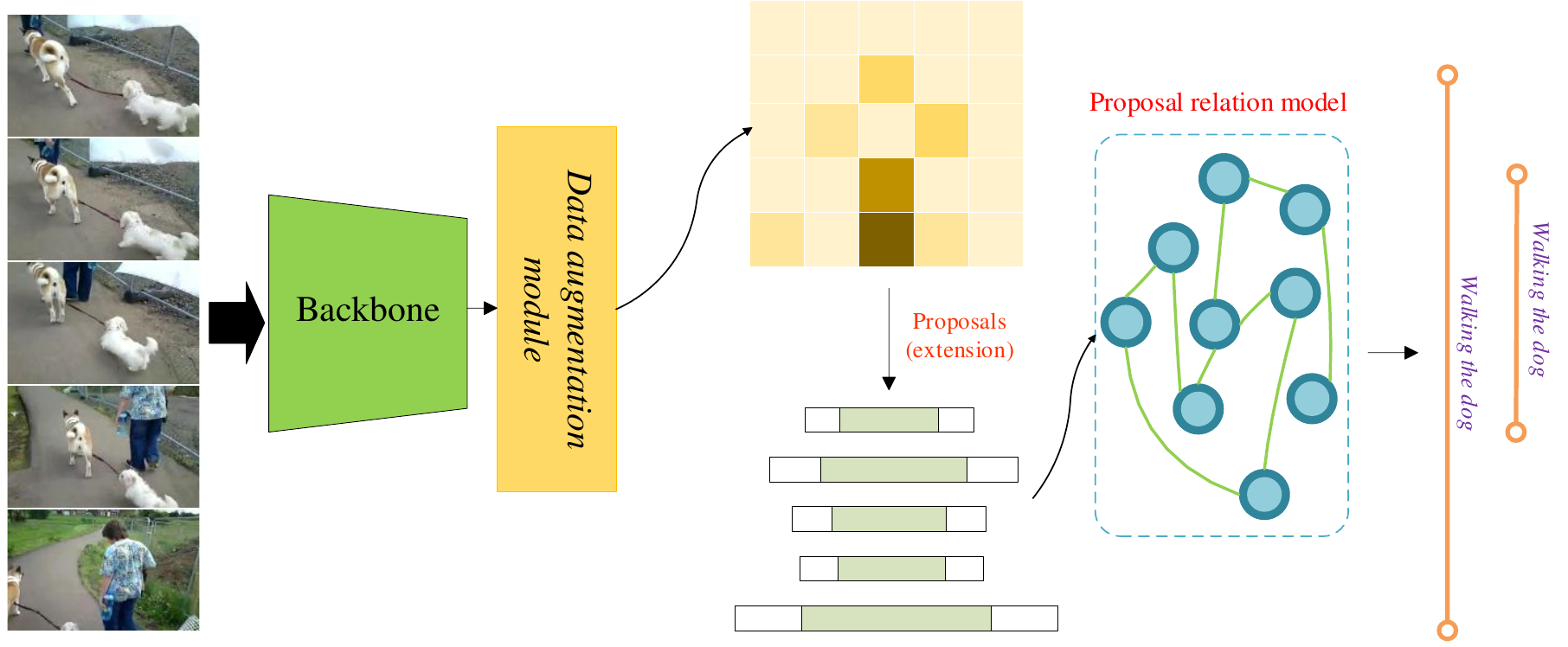}} 
\vspace{-1mm}
\caption{\label{figure-1} 
    The overall architecture of our solution. 
    Using a pretrained model, we extract the clip features for the input video, which are then fed into the data augmentation module. 
    Subsequently, we feed the enhanced features into a proposal generation network to obtain the confidence maps and then generate proposals. 
    Finally, the proposals are input into a proposal relation model to capture the relations, and output reliable and accurate proposals.
}
\vspace{-4mm}
\end{figure*}

%
\begin{figure}[t!]
\centering
\centering{\includegraphics[width=.8\linewidth]{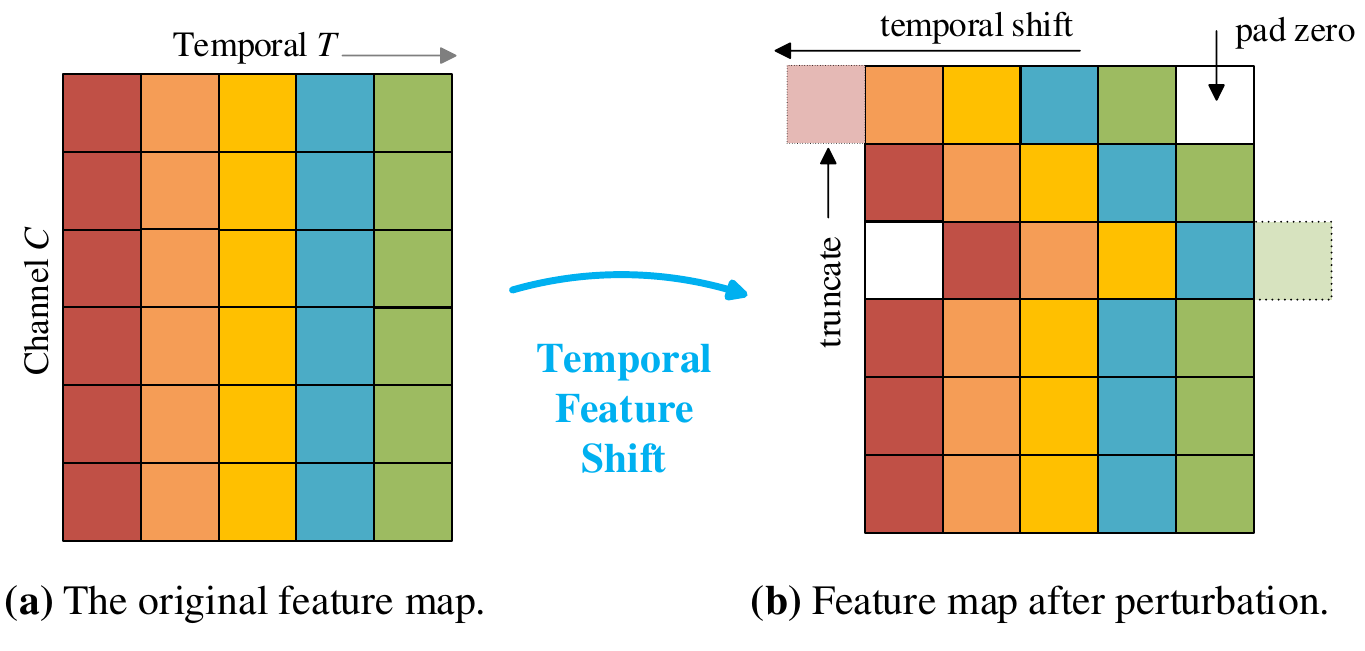}} 
\vspace{-1mm}
\caption{\label{figure-2} Detailed diagram of the \textit{data augmentation module}. Note that this module is borrowed from our SSTAP~\cite{wang2021self}.
}
\vspace{-4mm}
\end{figure}

\subsection{Channel-separated convolutional network}
Channel-Separated Convolutional Network (CSN)~\cite{csn} aims to reduce the parameters of 3D convolution, and extract useful information by finding important channels simultaneously.
It can efficiently learn feature representation through grouping convolution and channel interaction, and reach a good balance between effectiveness and efficiency.

\subsection{ViViT}
Due to transformer~\cite{transformer} has shown powerful abilities on various vision tasks, we apply the ViViT~\cite{VIVIT} as one of backbones.
ViViT is a pure Transformer based model for action recognition.
It extracts spatio-temporal tokens from input videos, and then encoded by series of Transformer layers.
In order to handle the long sequences of tokens encountered in videos, several efficient variants of ViViT decompose the spatial- and temporal-dimensions of the inputs.
We apply the ViViT-B/16x2 version with factorised encoder, which initialized from imagenet pretrained Vit~\cite{dosovitskiy2020image}, and then pretrain it on Kinetics700 dataset~\cite{k700}.
%


\subsection{Classification results}

Table~\ref{tab:classification} shows the action recognition results of the above methods on the validation set of ActivityNet v1.3 dataset~\cite{caba2015activitynet}. 
From the results, we can draw several following conclusions:
1) CSN model can outperform slowfast101 by 3.1\% with Kinetics400 pretraining on ActivityNet dataset ;
2) Transformer based model can indeed obtain better performance than CNN based models with 91.2\% Top1 accuracy. 
We then ensemble all the models and gain the champion result in ActivityNet 2020 with 1.8\%.

%
%

\begin{table}[t]
    \centering
\tablestyle{4pt}{1.5}
\small
\begin{tabular}{@{\;}@{\;}c@{\;}@{\;}|@{\;}@{\;}@{\;}c@{\;}|@{\;}@{\;}@{\;}c@{\;}|@{\;}@{\;}@{\;}c@{\;}}
     Model &  Pretrain  &  Top 1 Acc. &  Top5 Acc.\\
     \shline
    Slowfast101 & K400 &  87.1\% & 97.4\% \\
    \cline{1-4}
    Slowfast152 & K700 &  88.9\% & 97.8\% \\
    \cline{1-4}
    CSN & K400 &  90.3\% & 98.1\% \\
    \cline{1-4}
    ViViT-B/16x2 & K700 &  91.2\% & 98.0\% \\
    \cline{1-4}
    ANet 2020 champion & - &  91.8\% & 98.1\% \\
    \textbf{Ensemble} & \textbf{-} &  \textbf{93.6\%} & \textbf{98.5\%} \\
    
\end{tabular}\\
\vspace{+1mm}
    \caption{Action recognition results on the validation set of ActivityNet v1.3~\cite{caba2015activitynet}. K400 means pre-training on Kinetics 400~\cite{i3d}; K700 means pre-training on Kinetics 700~\cite{k700}. Note that we also use the ActivityNet 2020 champion results~\cite{wang2020cbr} for multi-model classification fusion.}
    \label{tab:classification}
\end{table}

    

\section{Proposal Relation Network}

In the section, we introduce our proposed PRN, as is shown in Figure~\ref{figure-1}.
PRN mainly contains two key modules: data augmentation module and proposal relation module. 
We will introduce each module in details below, and finally show the detection performance.

\subsection{Data augmentation module}

Temporal shift operation for action recognition is first applied in TSM~\cite{TSM}, and then applied as a kind of perturbations in  SSTAP~\cite{wang2021self} for semi-supervised learning. 
Here we reuse the perturbation as the feature augmentation. 
The temporal feature shift is a channel-shift pattern, including two operations such as forward movement and backward movement in the channel latitude of the feature map. 
This module can improve the robustness of the models.
%
%
The details are shown in Figure~\ref{figure-2}.

\subsection{Proposal relation module}

Recall that temporal action detection is to accurately locate the boundary of the target actions.
We explore the associations among proposals to capture the temporal relationships. 
The non-European space between the proposals makes it difficult to capture directly through the convolutional layer.
PGCN~\cite{pgcn} first employs GCN to model relations among proposals. 
However, PGCN suffers from multi-stage training and the adjacency matrix needs to be preset. 
Our PRN is an end-to-end trained framewoks, which brings in the power of self-attention~\cite{nonlocal} to capture interaction relationships among proposals.
Specifically, we use the attention mechanism on each proposal to obtain dependencies.

To evaluate proposal, we calculate AR under different Average Number of proposals (AN), termed AR@AN (\eg, AR@100), and calculate the Area under the AR \emph{vs.} AN curve (AUC) as metrics. 
Table~\ref{tab:proposal} presents the results of PRN and BMN on the validation set of ActivityNet v1.3, which prove that PRN can outperform BMN significantly. 
Especially, our method significantly improves AUC from 68.6\% to 69.3\% by
gaining 0.7\%.

\begin{table}[t]
    \centering
\tablestyle{4pt}{1.5}
\small
\begin{tabular}{@{\;}@{\;}c@{\;}@{\;}|@{\;}@{\;}@{\;}c@{\;}|@{\;}@{\;}@{\;}c@{\;}|@{\;}@{\;}@{\;}c@{\;}}
     Model &  Feature  &  AR@100 &  AUC \\
     \shline
    BMN & Slowfast101 &  75.8\% & 68.6\% \\
    \textbf{PRN} & \textbf{Slowfast101} &  \textbf{76.5\%} & \textbf{69.3\%} \\
    
\end{tabular}\\
\vspace{+1mm}
    \caption{Proposal performances on the validation set of ActivityNet v1.3.}
    \label{tab:proposal}
\end{table}

\begin{table}[t]
    \centering
\tablestyle{4pt}{1.5}
\small
\begin{tabular}{@{\;}@{\;}c@{\;}@{\;}|@{\;}@{\;}@{\;}c@{\;}|@{\;}@{\;}@{\;}c@{\;}|@{\;}@{\;}@{\;}c@{\;}}
     Model &  Feature  &  0.5 &  Average mAP\\
     \shline
    BMN & Slowfast101 &  56.3\% & 37.7\% \\
    PRN & Slowfast101 &  57.2\% & 38.8\% \\
    \cline{1-4}
    BMN & Slowfast152 &  55.5\% & 36.8\% \\
    PRN & Slowfast152 &  56.5\% & 38.0\% \\
    \cline{1-4}
    BMN & CSN &  56.9\% & 38.1\% \\
    PRN & CSN &  57.9\% & 39.4\% \\
    \cline{1-4}
    BMN & ViViT &  55.1\% & 36.7\% \\
    PRN & ViViT &  55.5\% & 37.5\% \\
    \cline{1-4}
    \textbf{Ensemble} & \textbf{-} &  \textbf{59.7\%} & \textbf{42.0\%} (\textcolor{blue}{\textbf{test: 44.7\%}}) \\
    
\end{tabular}\\
\vspace{+1mm}
    \caption{Action Detection results on the validation set of ActivityNet v1.3. Our PRN shows strong performance on multiple different features.}
    \label{tab:detection}
\end{table}

\subsection{Detection results}

We follow the ``proposal + classification" pipeline to generate the final detection results. 
Mean Average Precision (mAP) is adopted as the evaluation metric of temporal action detection task. 
Average mAP with IoU thresholds $[0.5 : 0.05 : 0.95]$ is applied for this challenge.

In order to demonstrate the effectiveness of PRN, we conduct experiments with different features, as is shown in Table~\ref{tab:detection}. 
The results shows that the proposed PRN can gain 1.3\% than BMN in terms of Average mAP. 
Then we ensemble all the results and reach 42.0\% on the validation set and 44.7\% on the test set.

Moreover, we can find that the Transformer based ViViT shows very strong performance on classification task but unsatisfactory on detection task when compared with the CNN models.
%
%
The reason may be that the Transformer tends to capture global information by self-attention operation, hence it loses local information which is also important for detection task.
%
%
Meanwhile, the models perform well on action task may not achieve better performance on the detection task.
Slowfast152 exceeds Slowfast101 by 0.8\% for classification, but suffers 1.8\% drop for detection.

%

\section{Conclusion}

In this report, we present our solution for temporal action detection task in ActivityNet Challenge 2021.
For this task, we propose a PRN to encode the relations among proposals, which contains data augmentation module and proposal relation module.
Experimental results show that PRN can outperform the baseline BMN significantly.
We also explore the ViViT model for the challenge, and experimentally show that it is more suitable for action classification than action detection.
By fusing all detection results with different backbones, we obtain 44.7\% Average mAP on the test set, which gains 1.9\% than the champion method in ActivityNet 2020.

\section{Acknowledgment}
This work is supported by the National Natural Science Foundation of China under grant 61871435 and the Fundamental Research Funds for the Central Universities no.2019kfyXKJC024.

{\small
\bibliographystyle{ieee_fullname}
\bibliography{egbib}
}

\end{document}